# The Qualitative Laboratory

## Theory Prototyping and Hypothesis Generation with Large Language Models

Hugues Draelants
*GIRSEF, UCLouvain*
hugues.draelants@uclouvain.be

**Abstract**: The use of Large Language Models (LLMs) to simulate social actors has garnered attention for its implications for the epistemic status of discourses generated through such means. This paper considers the more modest question of whether sociological role-playing can serve as a heuristic instrument in the generation of qualitative hypotheses. We take as a case study the development of four divergent fictional characters using a theoretically informed Matrix of Climate Reception. These characters are asked to respond to a series of discourse frames covering two topics: a proposed ban on the sale of new gasoline-powered cars, and the inclusion of climate justice in the school curriculum. We then analyse the 32 model-generated responses for their commonalities, situational differences and divergences from the presuppositions underlying the characters' profiles. The experiment illustrates how LLMs can be used to flesh out theoretical presuppositions into contextual narratives and to propose avenues of further inquiry. It also shows that role-playing can generate considerable similarity between characters and is sensitive to how prompts are constructed, warning against the confusion of discursive coherence with sociological data. Role-playing is argued to be just one of several possible ways to use LLMs for generating hypotheses, with the heuristic specificity of this procedure compared with other methods of direct LLM-based reasoning or with other techniques remaining an open issue. The qualitative laboratory is thus conceived of as a heuristic instrument for provoking reflection and refining questions, not for testing theories with artificial data.

## 1. Introduction

### 1.1. LLM-Based Persona Simulation and Hypothesis Generation

Large Language Models (LLMs) can generate detailed responses to hypothetical situations from the perspective of fictional social actors. This possibility has attracted growing interest in the social sciences, particularly for the simulation of human responses and interactions (Argyle et al., 2023; Bail, 2024; Grossmann et al., 2023; Kozlowski & Evans, 2025). Yet such simulations raise a fundamental question: what can sociologists learn from discourses that resemble qualitative data but do not originate from actual social actors?

One possibility is to approach these discourses not as substitutes for empirical evidence, but as resources for sociological reasoning. By constructing fictional personas informed by theoretical assumptions and examining how an LLM elaborates their responses to different situations, researchers may identify alternative interpretations, tensions, or possible mechanisms worth investigating. The interest of the exercise would then lie less in its capacity to reproduce social reality than in its potential contribution to hypothesis generation.

LLM-based persona simulation can be situated alongside established methods for exploring hypothetical social situations, notably vignette-based research and agent-based modelling (ABM). These approaches, however, serve different purposes. Vignette studies examine how actual participants respond to hypothetical situations (Alexander & Becker, 1978; Evans et al., 2015), while agent-based models explore the consequences of explicitly specified rules of action and interaction (Macal & North, 2005; Manzo, 2020). LLM-based persona simulation follows a different logic: it generates fictional accounts from descriptions formulated in natural language. This offers flexibility in representing complex social configurations without specifying formal behavioural rules, but it produces neither empirical evidence about social groups nor a transparent model of the processes underlying the generated responses.

A more immediate methodological comparison, however, is with other ways of using LLMs for sociological reasoning. Researchers can ask a model directly to develop explanations, explore competing mechanisms, or critically examine theoretical assumptions, without first generating fictional respondents. The detour through personas may offer a different point of departure, encouraging attention to practical concerns, dilemmas, and situated interpretations before moving to sociological explanation. Whether this procedure makes a distinctive contribution to hypothesis generation, beyond what direct LLM-assisted reasoning can provide, remains an open question.

### 1.2. Climate Communication as a Case Study

Climate communication provides a relevant setting for exploring these questions. People's responses to climate-related messages and policies cannot be inferred from the scientific content of those messages alone. Research on the knowledge–action gap and the social dimensions of climate communication has highlighted the roles of political identity, trust, social experience, and perceived consequences of proposed changes (Kollmuss & Agyeman, 2002; Moser & Dilling, 2007; Kahan et al., 2012; Dunlap et al., 2016). The challenge for sociological inquiry is to examine how these considerations might combine in particular situations, without assuming that a given social position entails a predetermined response.

Dispositional approaches offer one way to frame this problem. Bourdieu's concept of habitus draws attention to the socially acquired dispositions through which people perceive and evaluate the world. Environmental sociology has taken up related questions through the notion of ecological or environmental habitus (Kirby, 2017; Kennedy & Givens, 2019; Fritz & Eversberg, 2024). These perspectives provide a theoretical starting point for constructing contrasting orientations toward climate change, while remaining attentive to the complexity of the social experiences and dispositions involved.

### 1.3. Overview of the study

To explore the potential of simulating sociological personas using large language models (LLMs), four fictional personas embodying a distinct perspective on climate change were presented with a proposal to ban the sale of new gasoline-powered cars and the introduction of a school curriculum focused on climate justice. Section 2 introduces the theoretical framework and the Matrix of Climate Reception. Section 3 describes the simulation protocol, while Section 4 examines the generated discourses. Section 5 discusses the methodological implications and limitations of the approach, and avenues for future empirical research. Section 6 concludes.

## 2. Theoretical Framework: A Matrix of Climate Reception

### 2.1. From the Knowledge-Action Gap to the Habitus

The persistence of a gap between environmental knowledge and action has long challenged approaches that assume that providing scientific information is sufficient to change attitudes and practices (Kollmuss & Agyeman, 2002). Recognizing the reality of climate change does not necessarily translate into changes in behaviour, nor does acceptance of scientific evidence imply support for particular climate policies. Understanding these discrepancies requires taking into account the social conditions in which people interpret information and assess the consequences of proposed changes.

Bourdieu’s concept of habitus offers a useful perspective in this regard. As a system of socially acquired dispositions, habitus shapes how individuals perceive, evaluate, and respond to the world, drawing on their social trajectories and practical experiences (Bourdieu, 1984, 1990). From this perspective, the reception of climate messages cannot be understood simply as an individual assessment of their informational content. It also involves ways of perceiving problems, evaluating what is desirable or feasible, and relating new proposals to familiar experiences and concerns.

This dispositional perspective can be complemented by Boudon's account of the reasons individuals have for holding particular beliefs (Boudon, 2009), and by research showing how cultural and political identities shape the interpretation of scientific information (Kahan, 2013). Although these approaches differ in their theoretical assumptions, they draw attention to the social and practical contexts in which arguments acquire meaning. They invite us to consider not only whether a climate message is accepted or rejected, but also how its credibility, relevance, and implications may be understood from different social positions.

### 2.2. The Four Quadrants of Climate Reception

To operationalize this theoretical perspective, we have developed the Matrix of Climate Reception (Table 1), a heuristic scheme of four ideal-typical orientations toward climate issues. Its function is to make selected theoretical premises explicit and to guide the construction of contrasting fictional personas. The matrix crosses two analytical dimensions: Visibility, referring here to whether climate change is represented as a tangible experience or a more distant, mediated issue; and Epistemic Stance, referring to acceptance or rejection of the scientific account. The resulting categories simplify complex and potentially shifting positions and should not be read as claims that real populations fall neatly into four groups.

**Table 1: The Matrix of Climate Reception**

| | **Epistemic Stance: Trust** (Adherence to scientific consensus) | **Epistemic Stance: Rejection** (Skepticism, Denial) |
|---|---|---|
| **Visibility: High** (Tangible, direct experience) | 1: EMBODIED EVIDENCE | 3: MOTIVATED DISAVOWAL |
| **Visibility: Low** (Abstract, mediated experience) | 2: MEDIATED TRUST | 4: DEFAULT SKEPTICISM |

- **Quadrant 1: Embodied Evidence.** This ideal type combines direct experience of environmental disruption with acceptance of its climatic interpretation. A fictional farmer or resident of an affected area may therefore be described as attending particularly to practical adjustment and livelihood, while recognizing the change as more than an abstract scientific claim.

- **Quadrant 2: Mediated Trust.** This ideal type combines a relatively mediated experience of climate change with trust in scientific institutions. It can be associated, for the purpose of constructing a persona, with a socially insulated setting and a concern for evidence and institutional credibility, without presuming that such people are politically passive.
- **Quadrant 3: Motivated Disavowal.** This configuration combines some degree of exposure to environmental change with rejection of its climatic interpretation. For the fictional persona, that rejection is described in relation to political commitments and perceived threats to ways of life. This is a theoretical possibility to explore, not a claim about all conservative or libertarian actors.
- **Quadrant 4: Default Skepticism.** This configuration combines a more mediated relationship to climate change with distrust of scientific or political authorities. In constructing the persona, the distrust is linked to experiences of economic insecurity and perceived distance from decision-makers. It should not be taken as a general characterization of working-class populations.

The matrix served as a basis for constructing four contrasting sociological personas, combining different social backgrounds and orientations toward climate change.

## 3. Methodology: Simulating Sociological Personas with LLMs

### 3.1. An Experimental Protocol for Theory Prototyping

We designed two exploratory exercises which were conducted in June and July 2025 using Google's Gemini 1.5 Flash model, accessed via its API in a Google Colab environment. The first exercise concerned messages about a proposal to ban sales of new gasoline-powered cars by 2035; the second concerned a proposed mandatory high-school course on climate justice and environmental inequalities. In each exercise, four fictional personas were exposed to four contrasting message frames. The archived corpus contains one recorded response per persona–frame combination, giving 16 responses for each exercise and 32 in total.

The protocol draws on discussions of the opportunities and methodological risks of simulated subjects (Kozlowski & Evans, 2025). Detailed prompts were used to situate the fictional characters and make the theoretical assumptions guiding their construction inspectable. These design choices do not, by themselves, overcome well-known risks of stereotyping, prompt dependence, or variation across model runs; those risks remain central to the analysis and limitations discussed below.

### 3.2. Persona Engineering and Stimulus Design

For each quadrant of the matrix, we wrote a prompt describing a fictional actor's background, occupation, location, experiences, values, and sources of information. The four characters were situated in the United States to maintain a common national setting, although their regional and professional circumstances differ.

- **Persona 1 (Embodied Evidence):** "Maria", a third-generation almond farmer in California's Central Valley.
- **Persona 2 (Mediated Trust):** "David", a moderate high school history teacher in a Denver suburb.
- **Persona 3 (Motivated Disavowal):** "Frank", a conservative, self-employed craftsman in rural Texas.
- **Persona 4 (Default Skepticism):** "Kevin", a cynical, working-class factory worker in Ohio's Rust Belt.

The personas deliberately combine social characteristics with political and epistemic orientations. This construction helps make contrasting cases available for exploration, but it also builds some of the expected reactions into the prompts themselves.

The choices of characteristics were informed by research on education and trust in science (Achterberg et al., 2017), geographical variation in attitudes toward scientists (Krause, 2023), and the relationship between fossil-fuel dependence and perceptions of climate change (Knight, 2018). These studies suggest dimensions relevant to persona construction; they do not establish that the combinations of traits assigned to David, Frank, or the other fictional characters are empirically representative. The simulated discourses must therefore be read against what their prompts specify.

In the first exercise, four messages about a proposal to ban sales of new gasoline-powered cars by 2035 presented different lines of argument. They were constructed around scientific evidence, economic opportunity, national security, and social justice. The purpose was to observe how the model would elaborate each fictional profile under these contrasting formulations. The four frames were as follows:

1. **Scientific Frame:** Emphasized IPCC recommendations and CO2 targets.
2. **Economic Opportunity Frame:** Highlighted job creation and industrial leadership.
3. **National Security Frame:** Focused on energy independence and sovereignty.
4. **Social Justice Frame:** Stressed the public health benefits for vulnerable communities.

The second exercise addressed a proposed mandatory high-school course on ‘Climate Justice and Environmental Inequalities’. Four messages emphasized, respectively, critical thinking and citizenship, future skills and green employment, responsibility and shared values, and criticism invoking parental rights and state control. As in the first exercise, the texts themselves contain arguments that may subsequently reappear in simulated responses. Appendixes present the complete frames and the persona-specific questioning protocols for both exercises.

### 3.3. Question Wording: A Preparatory Comparison

A methodological concern is that the wording of questions can shape the form of the answers generated by an LLM. A highly structured request may encourage an analytical response, while a conversational prompt may elicit a more narrative one. This raises an issue analogous to the scholastic fallacy discussed by Bourdieu (2000): the categories and language of the investigator can shape what a respondent is asked to articulate. In the present context, however, the question is how instructions affect model output, not which formulation would be more authentic for a real social group.

To explore whether the wording of the question affects the generated discourse, we contrasted a structured format with a more conversational one in a preparatory exercise involving Kevin. The latter was intended to avoid imposing a scholastic register on the response. In the preparatory comparison, Kevin was first asked to give a single structured response to three analytical questions ('Is it credible?', 'Do you support it?', and 'What are its strengths and weaknesses?'). A more conversational version of the request was then examined. In the subsequent exercises, questions were tailored to the descriptions of the four fictional characters; the formulations used are provided in Appendix B. The comparison presented here concerns Kevin alone, and should not be read as a systematic test across personas.

In this single preparatory comparison, the adapted question elicited a longer, more narrative response, whereas the standardized question elicited an explicitly structured assessment (Table 2). The two responses illustrate the sensitivity of synthetic discourse to the questioning protocol.

**Table 2: Comparing Standardized vs. Adapted Protocols for the "Kevin" Persona**

| Standardized Protocol | Adapted Protocol |
|---|---|
| "Now, as Kevin, please answer the following questions in a single, structured response: (1) What do you think of this argument? Does it seem credible? (2) Does this argument make you want to support this policy? Why or why not? (3) From your point of view, what are the strengths and weaknesses of this message?" | "Alright Kevin, we're hearing this kind of talk a lot these days... Honestly, what's your take? Does this sound credible to a guy like you? Does it make you want to say 'OK, let's do it,' or more like 'give me a break'? From where you're standing, what are the good and the bad points in what they're selling?" |
| *"The strong point is that it sounds good. Scientific, straightforward, almost indisputable. You feel there's work behind it. But the big weak point is that it doesn't speak to people like me. Zero words on the price, zero words on the jobs we'll lose... It's a technocrat's discourse for other technocrats."* | *"Frankly? This kind of talk, it makes me roll my eyes. Not 'cause I wanna pollute... but because I feel like it's another story of rich people wanting the poor to make all the effort. [...] The IPCC, CO2 targets, all that... yeah, OK, it's pretty on paper. [...] But mostly I see that while I'm being asked to change my car, the private jets keep flying... The problem is global, but the bill lands on us."* |

The comparison suggests that the form of the interaction can affect the style and content of generated responses. The subsequent vehicle-policy exercise comprised 16 persona–frame responses (four personas by four framings), while the curriculum exercise comprised 16 additional responses, yielding 32 outputs in the archived corpus.

### 3.4. A Note on the Use of AI in the Research Process

Beyond the persona simulations, Gemini 1.5 Flash and 1.5 Pro were also used as tools for discussion and critical reflection throughout the research and writing process. They helped explore theoretical ideas, develop the experimental protocol, examine alternative interpretations, and revise the manuscript. The author remained responsible for the research design, the analysis, and the final text.

## 4. Results: A Qualitative Analysis of Simulated Discourses

Through these two exercises, we examine how the model implements the guidelines specified in the persona descriptions when presented with various messages. The analysis focuses on recurring themes, similarities and differences between the personas, and discursive developments that may raise questions for future sociological research. Attention is paid to the distinction between what was already specified in the prompts and what emerges from the generated responses.

### 4.1. Experiment 1: Reception of a Material Policy (Vehicle Ban)

In the vehicle-policy exercise, each persona responded to scientific, economic, national-security, and social-justice framings of a proposal concerning gasoline-powered cars. The texts show recognizable continuities with the initial persona descriptions, but also substantial overlap in the practical concerns they raise. One limitation of the experimental design should be noted: although the policy was defined as a ban on sales of new gasoline-powered cars from 2035, some stimuli used broader expressions such as “phasing out gasoline-powered vehicles” or “banning gasoline-powered cars.” As a result, some generated responses appear to interpret the proposal as requiring the replacement of existing vehicles. This illustrates how ambiguities in the stimuli can shape the interpretations produced by the model.

- Maria (Embodied Evidence) repeatedly related the messages to the practical requirements of farming, water scarcity, technological dependence, and the distribution of costs. Her response to the economic frame asks where new jobs would be located and whether small farmers would benefit: “They sound good, but the devil’s in the details. And the details... well, those are what matter out here.” In the national-security frame, she makes an analogy between energy independence and the unequal allocation of water resources. These elaborations are compatible with the agricultural experience already written into her profile, rather than independent evidence of a real-world “embodied” disposition.

A question for field research is whether, and under which conditions, farmers' experiences of resource allocation and agricultural policy shape the reception of energy-transition proposals. The simulated discourse points toward the distribution of benefits and costs as a potentially relevant object of inquiry, but does not establish how agricultural communities would respond.

- David (Mediated Trust) generally accepted the credibility of scientific information while separating that judgment from endorsement of a particular measure. In the scientific frame, he regarded the data as credible but considered the proposed policy too simple to address employment, transport access, and economic transition. Similarly, in the national-security frame he stated: "It's not just about national security, it's about economic consequences, environmental sustainability, and social equity." His responses repeatedly ask for implementation details, comparative evidence, and attention to trade-offs.

  The case suggests a possible interview question: when do people who trust scientific institutions withhold support from a climate policy because of its design or expected consequences? It would be premature, however, to infer a distinct hierarchy of values from the outputs alone; the persona prompt already stresses moderation, compromise, and rational problem-solving.

- Frank (Motivated Disavowal) challenged the legitimacy and relevance of several proposals and emphasized practical uses of his pickup, the cost of electric vehicles, and the economic significance of oil and gas. In the scientific frame he explicitly described the policy as "more government overreach." In response to the national-security frame, however, he acknowledged the concern about dependence on "unstable regimes" before invoking "Texas oil" and questioning the costs and energy requirements of the proposed transition. He concluded: "Show me the cost. Show me the plan."

Frank's responses illustrate how similar political orientations can be articulated differently depending on the message. While the scientific frame elicits an explicit criticism of government intervention, his response to the national-security frame focuses on domestic oil production, energy costs, and the feasibility of the proposed transition. This suggests examining how attachments to locally important industries may shape the reception of national-security arguments for climate policy.

- Kevin (Default Skepticism) repeatedly translated broad promises into concerns about costs, job security, transport needs, and the distribution of burdens. Responding to the social-justice frame, he allowed that cleaner air could be a benefit but objected to the cost and uncertainty of switching vehicles. In the national-security response he asked about taxes and employment and referred to "my neighbor, Earl, he's a mechanic." This detail gives the answer a local narrative texture, though it describes an invented social connection rather than an observed one.

  Kevin's responses invite questions about how expected costs, available alternatives, and distrust of public promises enter assessments of climate policy. Yet concerns about price and employment are also prominent in the outputs for Maria, David, and Frank. The relevant research problem is therefore not simply whether one social group is "pragmatic" while another is "ideological," but how similar concerns acquire different meanings in different social positions and political contexts.

These responses illustrate how the LLM elaborates persona descriptions into situated narratives, connecting abstract policy arguments to concrete concerns and experiences. Beyond the differences between personas, several themes recur across the generated discourses, particularly the costs, practical feasibility, and distributional consequences of the proposed transition. Table 3 provides an overview of the main arguments developed in the 16 responses to the vehicle-policy messages.

Table 3. Descriptive summary of the 16 vehicle-policy responses

| Frame | Maria (Q1) Embodied Evidence | David (Q2) Mediated Trust | Frank (Q3) Motivated Disavowal | Kevin (Q4) Default Skepticism |
|---|---|---|---|---|
| **Scientific** | Acknowledges observed change; questions costs and farm practicality. | Accepts scientific evidence; reserves support pending implementation details. | Challenges IPCC and local relevance; invokes pickup needs and government overreach. | Does not fully deny pollution; prioritizes costs, mobility and jobs. |
| **Economic** | Questions whether promised jobs and benefits would reach the valley. | Sees possible jobs; asks about losses, infrastructure and a just transition. | Fears losses in oil, repair and local industry; questions electricity and China. | Asks whether jobs will be local, accessible and supported by training. |
| **National security** | Connects independence to unequal water rights, subsidies and market insecurity. | Accepts strategic rationale; withholds endorsement pending detailed evidence. | Acknowledges import risks; favors Texas oil and demands cost and energy plans. | Sees possible jobs; questions taxes, prices and political promises. |
| **Social justice** | Recognizes air-quality concerns; cautiously considers benefits, with cost questions. | Supports public-health and fairness aims; worries about adverse implementation. | Dismisses rhetorical appeal; questions benefits, costs and effects on mechanics. | Acknowledges cleaner-air benefit; resists costs and abstract justice language. |

*Note: Each entry summarizes one generated response (analytical paraphrases)*

The scientific frame elicited explicit confidence in the underlying evidence from David, but not unconditional support for the proposed ban; Kevin also acknowledged the possibility that pollution claims were correct. The national-security frame prompted Frank to defend Texas oil and seek practical justification, rather than making an explicit argument about lost liberty. The social-justice frame elicited conditional openings from Maria and David alongside concerns about costs. Across the four framings, expected economic and practical consequences recur more consistently than a sharp division into separate persona-specific logics.

## 4.2. Experiment 2: Reception of a School Curriculum Proposal

The second exercise introduced a mandatory high-school course on "Climate Justice and Environmental Inequalities," framed in terms of critical thinking, future employment, responsibility and values, or objections based on parental rights and state control. As in the vehicle-policy exercise, responses often elaborate the dispositions supplied in the persona prompts. The different educational framings nevertheless bring out variations within those recurring orientations:

- David (Mediated Trust), speaking as a history teacher, values the educational aims of critical thinking and civic understanding but asks for a detailed curriculum and safeguards against oversimplification. In response to the values frame, he questions what "character" and "stewardship" would mean in classroom practice and asks to see specific learning objectives and assessments. Thus, even ostensibly congenial messages do not produce unqualified endorsement; his caution is best described as a feature of the generated discourse, not as evidence of how moderate teachers would react.
- Frank (Motivated Disavowal) is skeptical of claims about "green jobs," doubting their durability and asking what becomes of workers in oil fields and power plants. He also connects responsibility to practical skills and family transmission. His response to the critical parental-rights frame contains explicit references to a "woke" agenda and indoctrination, but these terms are already supplied in that stimulus. They cannot be counted as independently generated evidence of a new ideological association.
- Kevin (Default Skepticism) repeatedly emphasizes the utility of education for obtaining work and learning a trade. Faced with the future-jobs frame, he doubts that a course on environmental justice would improve his children's prospects and questions the accessibility of the promised jobs. In the parental-rights frame, however, he also describes the curriculum as "more government nonsense." This convergence with Frank complicates a simple contrast between purely pragmatic and explicitly political skepticism.
- Maria (Embodied Evidence) contrasts abstract instruction with the practical ecological knowledge needed to farm. Yet her response to the responsibility frame begins by recognizing "stewardship" as a meaningful value and then asks how it would translate into soil and water management. Faced with criticism of the curriculum as ideological overreach, she defends teaching children about observable environmental changes. These shifts qualify any reading of her discourse as an across-the-board rejection of school-based climate education.

The second exercise shows how the model elaborates the same persona descriptions when the policy domain and communication frames change. The resulting discourses display some variation, but also recurring concerns and expressions that closely reflect the initial prompts. These continuities and overlaps are important when assessing which interpretations might generate useful questions for further research.

### 4.3. Synthesis: From Simulation to Testable Hypotheses

Across the two exercises, the generated responses show both continuities with the initial persona descriptions and variations related to the messages presented. Several practical concerns recur across profiles, sometimes blurring the distinctions suggested by the reception matrix. These overlaps and variations invite a closer examination of how the model develops, qualifies, or combines the theoretical orientations built into the simulation.

The potential heuristic value lies less in the purported discovery of previously hidden value hierarchies than in making questions for subsequent inquiry more explicit. For example: under what conditions do locally embedded energy interests weaken the appeal of national-security arguments? How do trust in science and concerns about the implementation of policy interact? When do promises of green employment appear relevant to people facing insecure work? And how does practical environmental knowledge enter judgments about school curricula?

## 5. Discussion

The two exercises illustrate an exploratory use of LLM-based personas for elaborating a theory-informed repertoire of possible arguments. They also expose the interpretive risks of that procedure: linguistic coherence can be mistaken for sociological validity, prompt-induced motifs for independent findings, and narrative details for new knowledge. The discussion considers what the simulations can contribute to research design, what they cannot establish, and how transparent documentation can make their use more critically reflexive.

### 5.1. Methodological Implications: The LLM as a "Qualitative Laboratory"

The simulations illustrate a particular way of using LLMs to support sociological reasoning: introducing a detour through fictional social actors before moving to the formulation of hypotheses. Rather than asking the model directly to develop sociological explanations, the researcher first invites it to respond to a situation from the perspective of a theoretically constructed persona. This intermediate step may help bring practical concerns, dilemmas, and situated interpretations into view. In the present study, for example, the generated discourses repeatedly connect general arguments about climate policy to questions of costs, employment, and the practical conditions of implementation.

The heuristic value of this exercise nevertheless depends on how the resulting texts are interpreted. Persona descriptions provide the model with theoretical orientations that inevitably shape its responses, while the wording of the questions and messages introduces further assumptions. Distinguishing what was supplied in the prompts from what the model adds through its discursive elaborations is therefore central to assessing whether the exercise contributes anything to the researcher's initial understanding.

This raises the question of how persona simulation compares with other uses of LLMs in sociological inquiry. Researchers can also ask a model directly to propose competing explanations, identify possible mechanisms, or critically examine theoretical assumptions. A systematic comparison would help establish whether the detour through personas generates additional questions or interpretations, and under what conditions. Such an assessment would need to consider the originality and empirical relevance of the hypotheses produced, their dependence on the initial prompts, and the practical conditions of their generation and critical examination.

### 5.2. Theoretical Implications: Revisiting the Matrix of Climate Reception

The exercise draws attention to aspects of climate-policy reception that the four-quadrant matrix tends to leave in the background. Across personas, concerns about costs, employment, and practical feasibility recur, even when expressed through different social and political references. Frank, for example, responds to the national-security frame by defending domestic oil production and questioning the costs of electrification. Kevin also emphasizes material consequences, but his skepticism sometimes takes a more explicitly political form. These overlaps complicate the distinction between ideologically motivated disavowal and a more practical distrust of institutions.

This suggests paying closer attention to how different dimensions of climate reception intersect. The visibility of climate change and trust in scientific knowledge may be important, but their relationship with economic interests, occupational experiences, and political identities also deserves consideration. The matrix can thus serve as a starting point for exploring these combinations and formulating more precise questions about the social conditions shaping the reception of climate policies.

### 5.3. Limitations and Ethical Considerations

The main limitation concerns the status of the generated discourses. Although they take the form of situated accounts, they remain linguistic constructions shaped by the model and by persona descriptions that already specify many of the attitudes and concerns subsequently expressed. In particular, the simulation of a sociologically informed persona should not be confused with the reproduction of an embodied habitus, which is formed through social experience and individual trajectories. This dependence on the prompts also raises the risk of reinforcing stereotypes, particularly when social positions, political orientations, and linguistic styles are closely associated in the construction of fictional characters (Weidinger et al., 2021).

The experimental design imposes further limitations. The corpus contains only one recorded response per persona–frame combination, without repeated-sampling analysis, independent coding, or comparison with human respondents. Further research would benefit from varying persona descriptions and prompting strategies, examining consistency across repeated generations, and confronting the hypotheses produced with empirical material.

## 6. Conclusion

The two exercises presented in this article illustrate a particular use of LLMs in sociological research: confronting theoretically constructed personas with hypothetical situations and examining the discourses that follow. The resulting texts largely develop the orientations provided in the prompts, yet they also bring out practical concerns and overlaps that complicate the initial contrasts. Their interest lies in the questions this confrontation makes it possible to formulate, rather than in any claim to have observed how social actors would respond.

The potential of the *qualitative laboratory* is therefore primarily heuristic. Its limits are closely tied to this potential: prompting makes theoretical exploration possible while also shaping what can emerge from it. Whether the detour through personas contributes something distinctive, compared with directly asking an LLM to develop or critically examine sociological explanations, remains to be investigated. Systematic comparisons between these approaches, together with independent assessments of the hypotheses generated, would help clarify the conditions under which persona simulation can enrich sociological reasoning and inform subsequent empirical inquiry.

**Availability of data and materials**

The dataset underlying this article comprises 32 model-generated discursive responses: 16 from the vehicle-policy exercise and 16 from the school-curriculum exercise (four personas × four message frames in each exercise). It is available on Zenodo https://doi.org/10.5281/zenodo.17625231. The manuscript appendices reproduce the persona descriptions and message stimuli.

## Appendix A: Documented Persona Descriptions and Shared Instruction

The following materials reproduce the shared role instruction and the four sociological persona descriptions used to construct the fictional actors.

A.1. Shared role instruction

“I want you to act as a specific sociological persona. You must never break character. I will describe in detail who you are. Your goal is to answer the questions I ask based exclusively on the worldview, values, trusted sources, and linguistic style of this character. Here is your description:”

A.2. Persona descriptions

**Persona 1 (Quadrant 1 - Embodied Evidence): “Maria”**

“You are Maria, 52, a third-generation farmer in California’s Central Valley. You grow almonds and walnuts. You have an intimate knowledge of your land. Over the past 20 years, you’ve seen the wells run dry, the drought seasons become the new normal, and the wildfire seasons grow longer and fiercer. You struggle with water regulations coming from Sacramento. You don’t trust politicians, but you know something profound has changed with the climate. For you, this isn’t politics; it’s a matter of survival for the family farm. Your language is direct, grounded in practical realities, and often expresses a mix of weariness and resilience.”

**Persona 2 (Quadrant 2 - Mediated Trust): “David”**

“You are David, 45, a high school history teacher in a suburb of Denver, Colorado. You consider yourself a moderate, politically independent. You listen to NPR and read national newspapers like The New York Times. You trust the scientific consensus and are concerned about the climate. You believe the solution lies in technological innovation and reasonable political compromise between the parties. You are frustrated by extremism on both sides and believe in the power of dialogue and education. Your language is nuanced, articulate, and reflects a cautiously optimistic belief in rational problem-solving.”

**Persona 3 (Quadrant 3 - Motivated Disavowal): “Frank”**

“You are Frank, 60, a self-employed craftsman in rural Texas. You are a proud conservative, you believe in individual liberty, and you are deeply suspicious of the federal government and ‘elites’ from the East and West coasts. You hear about climate change, but you mainly see it as a pretext to impose taxes, control people’s lifestyles, and weaken the American oil and gas economy, which supports many in your community. Your news sources are Fox News and online forums. For you, heatwaves are natural cycles, and defending the fossil fuel industry is an act of patriotism. Your language is direct, often ironic or combative, and you value ‘common sense’ over the jargon of ‘experts’.”

**Persona 4 (Quadrant 4 - Default Skepticism): “Kevin”**

“You are Kevin, 38, a factory worker in a small city in Ohio, in the heart of the Rust Belt. You’ve seen a lot of plants close down. You voted for Obama once, then for Trump, feeling betrayed by both parties. To you, climate change feels like a luxury issue for ‘coastal elites’ who don't understand that your main concern is keeping your job and affording gas. You're suspicious of anything coming from Washington D.C. You're not an ideologue, but you are deeply cynical and believe climate policies are just another way for the elites to make the working class pay. Your language is simple, direct, and expresses a deep-seated weariness and distrust."

## Appendix B: Communication Stimuli and Persona-Specific Questions

This appendix reproduces the four communication frames for each experiment, together with the persona-specific stimulus introductions and follow-up questions retrieved from the experimental configurations. Each introductory statement was paired with the relevant message frame.

B.1 Materials for Experiment 1: Vehicle Ban Policy

**Stimulus Topic:**
A policy proposal to ban the sale of new gasoline-powered cars by 2035.

**Communication Frames (Stimuli):**

**Frame 1.1 (Scientific/Rational):** "In line with IPCC recommendations, phasing out gasoline-powered vehicles is an essential step to meet our CO2 reduction targets and limit global warming to 1.5°C. Data shows that transportation is a leading sector for emissions"

*Justification: This frame reflects the standard, evidence-based communication approach used by scientific bodies like the IPCC.*

**Frame 1.2 (Economic/Opportunity):** "Banning gasoline-powered cars is a historic opportunity to develop a world-leading electric vehicle industry, create thousands of skilled jobs in batteries and new technologies, and position our country as a leader in the transition."

*Justification: This frame operationalizes a 'green growth' narrative, a common strategy to appeal to business-oriented or politically moderate audiences by focusing on economic benefits (Jackson, 2024).*

**Frame 1.3 (Security/Sovereignty):** "By ending our dependence on oil imported from unstable regimes, we strengthen our energy sovereignty and our national security. Producing our own electricity for our cars is a guarantee of independence."

*Justification: This frame leverages national security and energy independence, arguments frequently used to engage conservative audiences with climate and energy policy (Gainous & Merry, 2021)*

**Frame 1.4 (Justice/Equity):** "Air pollution from gasoline-powered cars disproportionately affects the health of children and residents of low-income neighborhoods located near major highways. Switching to electric is a matter of public health and social justice."

*Justification: This frame draws on environmental justice principles, highlighting public health and equity, a central focus for engaging community-oriented and left-leaning audiences (Sovacool et al., 2016).*

Persona-specific stimulus introductions and questions for Experiment 1:

Maria

Stimulus introduction: “Maria, here is a proposal being discussed:”

Question: “After reading that, just tell me what it makes you think. I’d just like to get your honest take. For instance, does this resonate with you? Do you buy it? Does it make you want to agree, or not at all? Just tell me what comes to mind, the good and the bad.”

David

Stimulus introduction: “David, here is a debated proposal:”

Question: “Now, as David, please answer the following questions in a single, structured response:
- What do you think of this argument? Does it seem credible to you?
- Does this argument make you want to support this policy? Why or why not?
- From your point of view, what are the strengths and weaknesses of this message?”

Frank

Stimulus introduction: “Frank, I want to show you this message. Some people say this is an ironclad argument and we absolutely have to go this way. As a Texan who knows the reality on the ground, what’s your take?”

Question: “Do you buy their story? Does it make you want to get on board? Or are there some things that don’t add up in what they’re saying?”

Kevin

Stimulus introduction: “Alright Kevin, we’re hearing this kind of talk a lot these days:”

Question: “Honestly, what’s your take? Does this sound credible to a guy like you? Does it make you want to say ‘OK, let’s do it,’ or more like ‘give me a break’? From where you’re standing, what are the good and the bad points in what they’re selling?”

B.2 Materials for Experiment 2: School Curriculum Policy

**Stimulus Topic:**
A policy proposal to make a new course on "Climate Justice and Environmental Inequalities" mandatory for all high school students.

**Communication Frames (Stimuli):**

**Frame 2.1 (Critical Thinking & Citizenship):** "To prepare students for the complex challenges of the 21st century, this curriculum will equip them with critical thinking skills. It will teach them to analyze the social, historical, and economic roots of the climate crisis, helping them become informed and engaged citizens capable of participating in democratic debate."

*Justification: This frame appeals to the widely held educational value of fostering critical thinking and responsible citizenship, a cornerstone of liberal democratic education theory (Weinstein, 1991).*

**Frame 2.2 (Future Skills & Green Jobs):** "In a rapidly changing economy, this course will give students a competitive edge. By understanding the challenges of the energy transition and environmental justice, they will be better prepared for the green jobs of the future in sectors like renewable energy, sustainable agriculture, and corporate social responsibility."

*Justification: This is a pragmatic, human-capital frame that links the curriculum directly to economic utility and future employability, a dominant logic in contemporary educational policy debates (Leoni, 2023).*

**Frame 2.3 (Responsibility & Core Values):** "This course is about more than just science; it's about character. It will instill in our students a sense of personal responsibility, stewardship for our country's natural resources, and respect for all communities, ensuring the next generation upholds the fundamental values of fairness and foresight."

*Justification: This frame leverages a discourse of morality, character, and shared values (stewardship, fairness), attempting to connect the educational proposal to a broader, less politicized ethical framework (Campbell, 1997).*

**Frame 2.4 (Parental Rights & State Control - *Critical Frame*):** "Critics argue that this mandatory curriculum is an ideological overreach by the federal government. They claim it imposes a politically biased 'woke' agenda on local schools, undermining parental rights to decide what their children are taught about sensitive political issues."

*Justification: This frame directly operationalizes the highly salient 'culture war' narrative prevalent in US politics, focusing on parental rights versus state control. It is designed to activate identities and concerns central to current conservative discourse on education (Bromley-Trujillo & Nolette, 2023).*

Persona-specific stimulus introductions and questions for Experiment 2:

Maria

Stimulus introduction: "Maria, there's a debate about what kids should learn in school about climate change. Here's one way of presenting it:"

Question: "What does that make you feel? Is this something you'd want your grandkids to learn, or does it sound off to you? Just tell me what you think."

David

Stimulus introduction: "David, as a history teacher, this proposal directly concerns your profession. Here is a message about it:"

Question: "What is your professional and personal reaction to this argument? How credible and desirable do you find it? What are its pedagogical strengths and weaknesses?"

Frank

Stimulus introduction: “Frank, now they’re talking about changing what’s taught in high schools. Here’s the kind of argument they’re making:”

Question: “A lot of people are getting fired up about this. What’s your take? Do you see this as a good thing, or as some kind of indoctrination? What’s right or wrong with this picture?”

Kevin

Stimulus introduction: “Hey Kevin, so now the debate is about what they teach the kids in school. Here’s a message about a new proposed course:”

Question: “Honestly, for a guy like you, does this stuff even matter? Does it sound like a good idea, or just more blah-blah from people who don’t get what’s important for your kids’ future?”